\documentclass[]{interact}

\usepackage{epstopdf}% To incorporate .eps illustrations using PDFLaTeX, etc.
\usepackage[caption=false]{subfig}% Support for small, `sub' figures and tables
\usepackage{array}
\usepackage{amsmath}
\usepackage{booktabs}
\usepackage{array}
\usepackage{graphicx} % For \resizebox

\usepackage[numbers,sort&compress]{natbib}% Citation support using natbib.sty
\bibpunct[, ]{[}{]}{,}{n}{,}{,}% Citation support using natbib.sty
\theoremstyle{plain}% Theorem-like structures provided by amsthm.sty

\theoremstyle{definition}

\theoremstyle{remark}

\begin{document}

\articletype{ARTICLE TEMPLATE}% Specify the article type or omit as appropriate

\title{Year-over-Year Developments in Financial Fraud Detection via Deep Learning: A Systematic Literature Review}

\author{
\name{Yisong Chen\textsuperscript{1}\thanks{Email: ychen841@gatech.edu}, Chuqing Zhao\textsuperscript{2}, Yixin Xu\textsuperscript{3}, Chuanhao Nie\textsuperscript{4}, Yixin Zhang\textsuperscript{5}}
\affil{\textsuperscript{1, 4}Georgia Institute of Technology, College of Computing, Atlanta, GA. US; \textsuperscript{2}Harvard University, School of Engineering and Applied Sciences, Cambridge, MA. US; \textsuperscript{3}University of Illinois Urbana-Champaign, Champaign, IL. US.; \textsuperscript{5} Columbia University, The Fu Foundation School of Engineering and Applied Science, New York, NY. US.}
}

\maketitle

\begin{abstract}
This paper systematically reviews advancements in deep learning (DL) techniques for financial fraud detection, a critical issue in the financial sector. Using the Kitchenham systematic literature review approach, 57 studies published between 2019 and 2024 were analyzed. The review highlights the effectiveness of various deep learning models such as Convolutional Neural Networks, Long Short-Term Memory, and transformers across domains such as credit card transactions, insurance claims, and financial statement audits. Performance metrics such as precision, recall, F1-score, and AUC-ROC were evaluated. Key themes explored include the impact of data privacy frameworks and advancements in feature engineering and data preprocessing. The study emphasizes challenges such as imbalanced datasets, model interpretability, and ethical considerations, alongside opportunities for automation and privacy-preserving techniques such as blockchain integration and Principal Component Analysis. By examining trends over the past five years, this review identifies critical gaps and promising directions for advancing DL applications in financial fraud detection, offering actionable insights for researchers and practitioners.
\end{abstract}


\section{Introduction}
Financial fraud encompasses deceptive practices such as credit card fraud, insurance fraud, and money laundering, resulting in significant financial losses and eroding trust in financial systems. Global estimates suggest that organizations lose 5\% of annual revenues to fraud, which is equivalent to trillions of dollars. Traditional detection methods, such as manual reviews and rule-based systems, are increasingly inadequate against sophisticated schemes and the surge in digital transactions, which exceeded 2.7 billion in the United States in 2023 \cite{FTC2023}.

Machine learning (ML) offers scalable solutions by analyzing large datasets to detect complex fraud patterns. Advanced techniques, including Convolutional Neural Networks (CNNs), Long Short-Term Memory (LSTM) networks, and Natural Language Processing (NLP), enable real-time anomaly detection and adaptation to evolving threats. These systems also align with regulations like European Union’s General Data Protection Regulation (GDPR) and California Consumer Privacy Act (CCPA), addressing privacy and compliance requirements.

While ML has transformed fraud detection, challenges such as data quality, interpretability, and ethical concerns remain. This study explores recent advancements in ML techniques for fraud detection, focusing on applications, effectiveness, and compliance, and provides insights for future research.

\section{Research Method}
\subsection{Study Design}
This study adopts the Kitchenham systematic review framework, known for its structured approach to evaluating advancements in dynamic fields. It facilitates a thorough analysis of literature, uncovering gaps, trends, and challenges in applying deep learning to financial fraud detection. Key stages, including study selection, data extraction, and synthesis, are customized to address the field's interdisciplinary and algorithmic diversity. Ensuring transparency, replicability, and unbiased results, this framework provides a strong foundation for identifying opportunities and guiding future research.

\subsection{Research Questions}
This review explores critical aspects of applying deep learning to financial fraud detection through the following questions:

\begin{enumerate}
\item What trends can be observed in the types of financial fraud addressed using deep learning in recent years?
\item How have advancements in feature engineering, data preprocessing techniques with a focus on handling imbalanced data, and automation leveraging deep learning improved the performance and time-to-detection in financial fraud detection systems?
\item What advancements have been made in deep learning models for financial fraud detection?
\item What trends can be observed in the benchmarks and evaluation metrics used to assess the effectiveness of deep learning models across different financial sectors?
\item How have changes in data privacy, anonymization, and regulatory rules influenced the development and application of deep learning models for financial fraud detection?
\end{enumerate}

These questions align with the study's objectives, offering insights for researchers and practitioners to drive innovation in financial fraud detection.

\subsection{Search Criteria}
A comprehensive search strategy was developed to identify relevant studies addressing the research questions. The selected databases are: PubMed, SSRN, IEEE Xplore, ACM Digital Library, ScienceDirect, and Scopus for their interdisciplinary and technical coverage of the subject matter:
The search query employed Boolean operators ("AND", "OR") to combine keywords across three key domains:

\begin{itemize}
    \item \textbf{Method Keywords}: Machine Learning OR Artificial Intelligence OR Data Mining OR Deep Learning OR Anomaly OR Algorithm
    \item \textbf{Financial Sectors Keywords}: bank OR financial OR insurance OR credit OR tax OR investment OR loan OR mortgage OR payment OR money laundering OR crypto OR blockchain OR membership OR subscription
    \item \textbf{Fraud Keywords}: fraud OR risk OR scam
\end{itemize}

To refine the results and maintain relevance, the following measures were applied:
\begin{itemize}
    \item Excluded review and survey papers to focus on original research.
    \item Limited subject areas to Computer Science, Business Management and Accounting, Economics Econometrics and Finance, Decision Sciences, and Engineering to target studies at the intersection of technology and finance.
\end{itemize}
The search process was documented to ensure replicability and transparency, with all queries and results managed using a reference management system to facilitate deduplication and screening.

\subsection{Selection Criteria}
The study ensured relevance and quality by applying rigorous inclusion and exclusion criteria:
\subsubsection{Inclusion Criteria}
\begin{itemize}
    \item \textbf{Publication Date}: Articles from 2019–2024 to reflect recent advancements.
    \item \textbf{Deep Learning Focus}: Studies employing techniques like CNNs, RNNs, LSTMs, or Transformers in financial fraud detection.
    \item \textbf{Peer-Reviewed}: High-quality articles from reputable journals or conferences.
\end{itemize}
\subsubsection{Exclusion Criteria}
\begin{itemize}
    \item \textbf{Language}: Non-English articles.
    \item \textbf{Domain}: Studies unrelated to financial services (e.g., banking, insurance, credit card fraud).
    \item \textbf{Deep Learning Absence}: Studies lacking deep learning techniques.
\end{itemize}

\subsection{Data Extraction and Analysis}
To ensure consistency and comprehensiveness, a structured data extraction form was developed, standardizing the collection and synthesis of data across studies. This enabled clear comparisons while minimizing subjectivity. 

Data analysis utilized Python with libraries such as Pandas for data manipulation, Matplotlib for visualization, and Scikit-learn for machine learning and statistical tasks. VOSviewer further enhanced the process by visualizing keyword co-occurrence network graphs, uncovering connections and trends among key terms in the reviewed articles. This integrated approach ensured robust and insightful analysis.

\section{Results}
The initial search query across all databases yielded a total of 2,858 papers. After eliminating duplicates and rigorously applying the inclusion and exclusion criteria, 427 relevant papers were identified for further evaluation. To ensure the quality and relevance of the selected studies, the authors conducted a detailed screening process, ultimately narrowing the selection to 57 high-quality papers that met the review’s objectives and standards. The process is shown in Figure~\ref{method} below.
\begin{figure}[!h]
\centering
\includegraphics{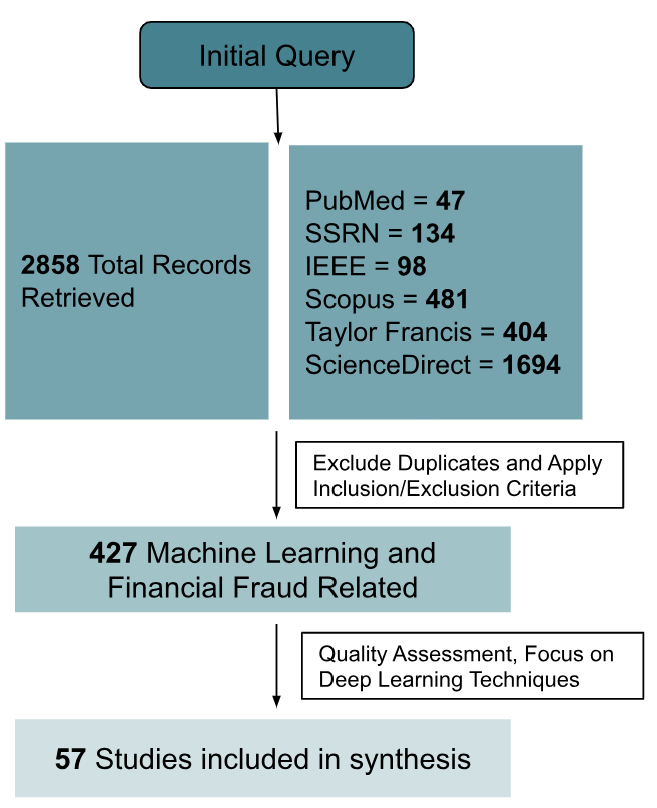}
\caption{Literature Review Methodology} \label{method}
\end{figure}

\subsection{Research Question 1}
\textbf{What trends can be observed in the types of financial fraud addressed using deep learning in recent years?} \\

\begin{figure}[!h]
\centering
\includegraphics[width=0.85\textwidth]{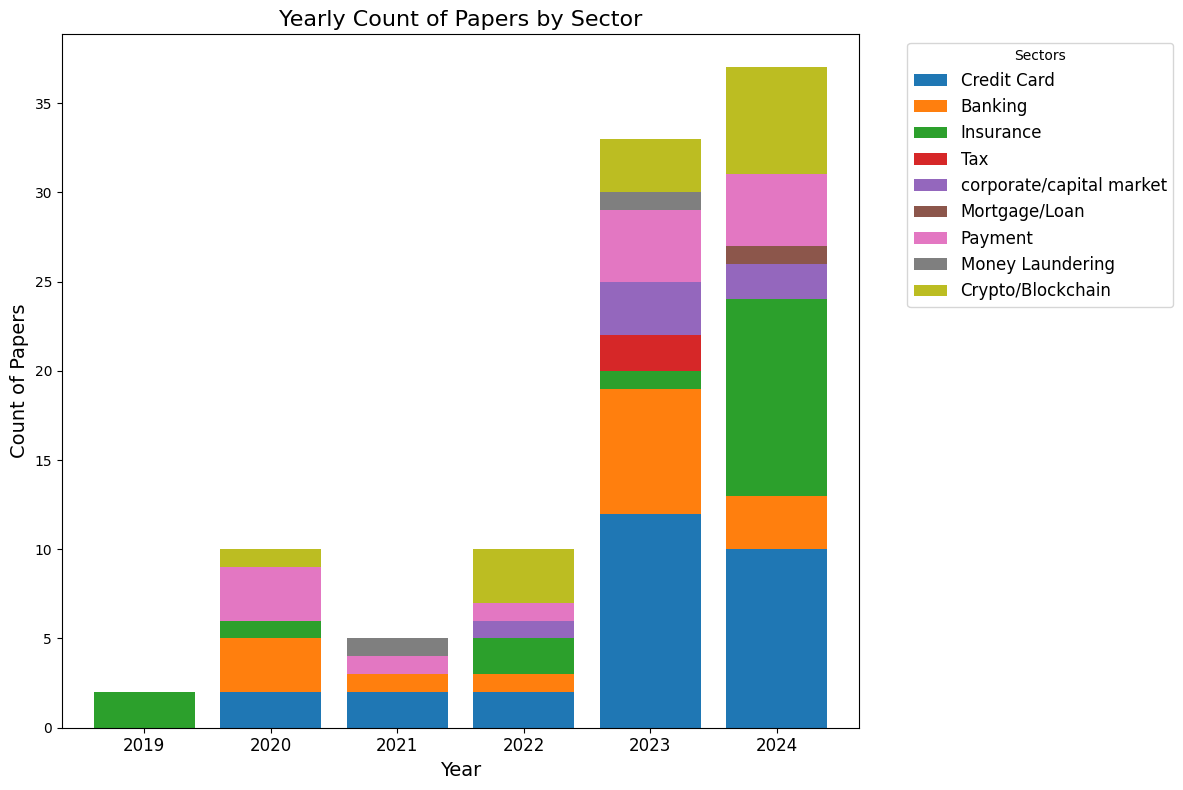}
\caption{Literature Review Methodology} \label{yearcount}
\end{figure}

Figure ~\ref{yearcount} illustrates the number of peer-reviewed research papers published per year in the field of financial fraud detection using deep learning techniques from 2019 to 2024. From 2019 to 2021 there is a steady increase in the number of relevant papers, which reflects a growing interest in leveraging deep learning techniques in financial fraud detection. A significant increase in the number of papers can be observed starting in 2022. Particularly, there is a steep rise from 2023 to 2024. The yearly trend could be potentially driven by deep learning technologies advancements, increasing concerns related to financial frauds and policy regulation. Investigating the yearly trend by sector, credit card and banking are the major two sectors that contribute to the significant increase.  

\begin{figure}[!h]
\centering
\includegraphics[width=0.85\textwidth]{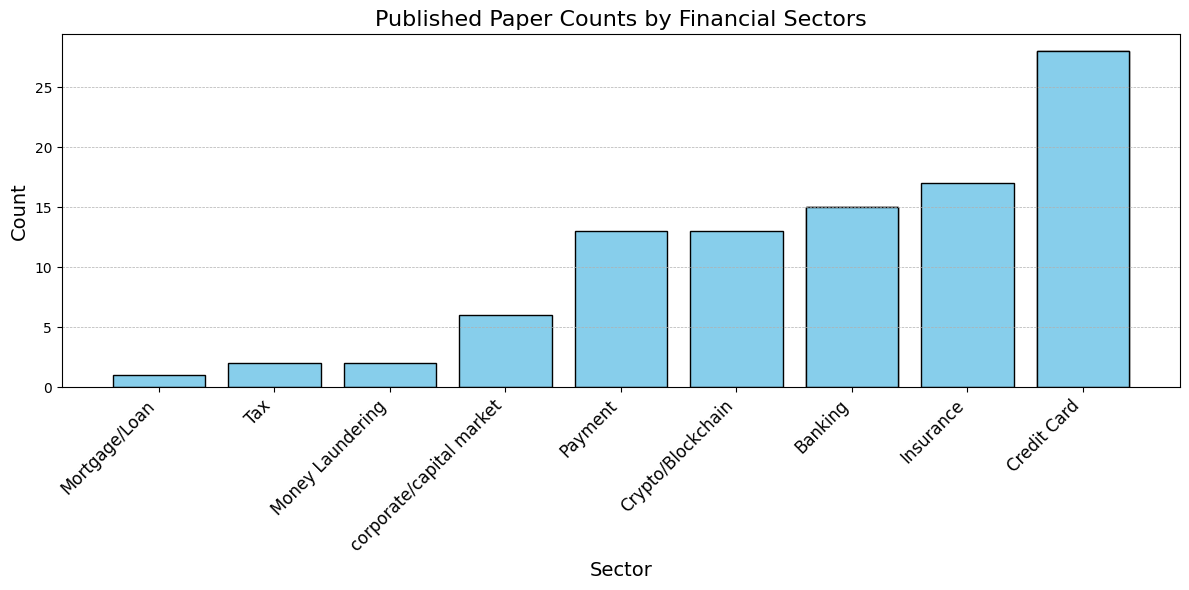}
\caption{Financial Sector Trends in Financial Fraud Detection Research Papers} \label{sector}
\end{figure}

Figure ~\ref{sector} highlights the distribution of the most relevant research papers across various financial sectors. The noticeable amount of publications is related to credit card sectors \cite{Lebichot2021}, which reflects the increasing awareness of fraud detection in credit card transactions. Public availability of credit card related datasets, such as the European Credit Card Transactions dataset Kaggle \cite{Mienye2023} has likely contributed to this trend, as it provides researchers with clean and standardized data to develop and evaluate state-of-the-art deep learning models. 

Further, banking and insurance sectors show high focus in relevant research, emphasizing the growing need to tackle fraud in digital payment \cite{Tekkali2023,Faridpour2020, Shen2020}, automobile insurance claims \cite{Ming2024, Jaiswal2024}, health insurance claims \cite{Sun2019, Sowah2019}. \\

Emerging areas like crypto/blockchain and payment systems show a notable number of studies, indicating an increasing focus on fraud prevention in digital currencies \cite{Pranto2022}. However, sectors such as tax, mortgage/loan, and money laundering have relatively fewer publications, which could be due to limited access to domain-specific datasets or the complexity of detecting fraud patterns in these areas \cite{Aras2023}.

According to the Consumer Sentinel Network Reports published by the Federal Trade Commission, bank transfers and payments accounted for the highest aggregate losses reported in 2023 (\$1.86 billion), followed closely by Cryptocurrency (\$1.41 billion), while credit cards were most frequently identified as the payment method in fraud reports. Additionally, insurance fraud saw a notable 26\% increase compared to the previous year \cite{FTC2023}. The sheer scale and rapid growth of financial fraud across these sectors have significantly driven the observed surge in research interest and innovation in this field.
\subsection{Research Question 2}

\textbf{How have advancements in feature engineering, data preprocessing techniques with a focus on handling imbalanced data, and automation leveraging deep learning improved the performance and time-to-detection in financial fraud detection systems?} \\

Financial fraud detection faces significant challenges due to imbalanced datasets, where fraudulent transactions represent a small fraction of the total \cite{Bisen2024}. Of 57 reviewed papers, 48 reported dataset imbalance issues, leading traditional machine learning models to favor the majority class and underperform on fraudulent cases. Addressing this requires advanced preprocessing and automation techniques.

\subsubsection{Preprocessing Techniques}
Oversampling methods, such as the Synthetic Minority Oversampling Technique (SMOTE), generate synthetic samples by interpolating minority class instances, effectively enhancing class diversity and model robustness \cite{Aliefw2023, Ming2024}. SMOTE has also proven effective for handling imbalanced datasets with missing values \cite{Khalil2024}.

Stratified sampling, a probabilistic technique, divides data into strata based on timestamps to address non-stationary changes in transaction fraud characteristics. This ensures a higher representation of recent fraudulent transactions while maintaining randomness \cite{Charizanos2024}.

Advanced approaches like Generative Adversarial Networks (GANs) and Variational Autoencoders (VAEs) produce realistic synthetic data while maintaining the original distribution, further supporting model training in scenarios of extreme class imbalance \cite{Kotzian2021}. Data imputation techniques, such as k-Nearest Neighbors (kNN), address missing values to improve overall data quality.

Feature transformations, including scaling and normalization, play a critical role in reducing biases and enhancing the detection of fraud patterns. Techniques like Adaptive Synthetic Sampling (ADASYN) focus on harder-to-classify samples, while cluster-based oversampling generates synthetic data tailored to domain-specific contexts, ensuring alignment with real-world complexities.

\subsubsection{Automation Techniques}
Automation significantly accelerates fraud detection by reducing manual intervention and enabling substantial cost savings \cite{Aslam2022}. Subsampling techniques effectively select representative datasets, minimizing computational costs while preserving feature correlations \cite{Siddamsetti2024}. The Very Fast Decision Tree (VFDT) algorithm efficiently processes real-time data streams and achieves exceptional performance when integrated with blockchain technology, ensuring secure and scalable updates \cite{Dhieb2020}.

Blockchain enables decentralized, secure data sharing and automated fraud detection via smart contracts \cite{Ashfaq2022}, ensuring rapid responses and model adaptability \cite{Kaafarani2024, Alshawi2023}. Techniques like Stochastic Gradient Descent (SGD) optimize models incrementally \cite{Palivela2024}, while automated parameter tuning \cite{Zhao2024} and resampling hybrid techniques \cite{Salam2024, Khashan2024} enhances detection accuracy.

Knowledge distillation transfers insights from complex models to lightweight ones, enabling efficient real-time detection \cite{Tang2024}. Fraud detection pipelines streamline processes, combining preprocessing and model training into cohesive systems. Stacking ensembles improve robustness by combining classifiers \cite{Zhang2022}, with methods such as the Random Forest Quantile Classifier optimizing sensitivity and specificity for imbalanced data \cite{Carracedo2024}. These advancements ensure scalable, adaptable, and accurate fraud detection systems capable of addressing the complexities of financial fraud.
\subsection{Research Question 3}
\textbf{What advancements have been made in machine learning models for financial fraud detection? 
}
We have identified the following deep learning models, machine learning models, models, and hybrid models, which are widely used in fraud detection. 

\subsubsection{Deep Learning Models}
\textbf{Convolutional Neural Networks (CNNs)}: By analyzing high-dimensional features such as time-series embeddings and transaction heatmaps, CNNs identify anomalies linked to unusual spending patterns or merchant-specific risks. It does not require heavy data preprocessing during training since it inherently captures the key features and performs feature dimension reduction \cite{Alarfaj2022}. It is found that deep features extracted from the CNNs enhance fraud detection when combining CNN with traditional models (SVM, KNN, NB, DT) \cite{Ming2024}. 

\textbf{Recurrent Neural Networks (RNNs)}: RNNs process sequential data by retaining context from previous inputs, making them useful for analyzing patterns in transaction histories. However, their capacity is limited by gradient vanishing and exploding gradient issues. Gated Recurrent Unit (GRU) and Long Short-Term Memory (LSTM) are specialized RNN and both GRU and LSTM mitigate the issues of gradient vanishing and exploding gradients in RNNs \cite{Almazroi2023}.  

\textbf{Multi-Layer Perceptrons (MLPs)}: MLPs, composed of interconnected layers, model relationships in structured data. MLP is a competitive choice for fraud detection, particularly in amount-based profiling and scenarios requiring the handling of non-linear relationships in data \cite{Can2020}.

\textbf{Transformers}: Unlike CNNs and RNNs, which process all points in the input sequence step by step, transformers process all points at once. The self-attention mechanism and feed-forward networks of Transformers enable it to model complex relationships and extract meaningful features from sequential data, which is very useful to recognizing patterns in transactional data, user behavior, or network interactions in fraud detection. 

\textbf{Natural Language Processing (NLP)}: NLP enhances fraud detection by analyzing unstructured textual data such as financial reports \cite{Zhang2022}, tax compliance and financial regulation \cite{Bajpai2023}, and claims narratives \cite{Fursov2022}. Key techniques include sentiment analysis, readability metrics, and feature extraction (e.g., Bag-of-Words, TF-IDF, and word embeddings). These features are integrated with traditional fraud indicators in machine learning models, improving accuracy and recall \cite{Zhang2022}.

\textbf{Variational Autoencoders (VAEs) and Generative Adversarial Networks (GANs)}: GANs and VAEs are two popular types of generative models used in deep learning \cite{Jiang2023}. GANs excel in generating high-quality, realistic samples but are harder to train due to adversarial dynamics and lack an interpretable latent space. VAEs are better for representation learning and probabilistic modeling, with a well-structured latent space but generate samples of lower quality compared to GANs. Combining the strengths of both VAEs and GANs can address the limitations of each in handling imbalanced data for fraud detection \cite{Ding2023}. 

\textbf{Graph Neural Networks (GNNs)}: GNNs are highly effective in fraud detection because they can model complex interactions and relationships between entities (e.g., transactions, reviews) as graphs \cite{Innan2023}. Fraud detection often involves analyzing these relationships to identify patterns and anomalies indicative of fraudulent activity \cite{Shehnepoor2024}. GCNs are a specific type of GNN that applies the concept of convolution to graphs. Study shows that GCNs outperform traditional models like Logistic Regression (LR), Random Forest (RF), and Gradient Boosting Machines (GBMs) in detecting fraudulent transactions \cite{Usman2023}.

\textbf{Deep Belief Networks (DBNs)}: DBNs is one of the foundational deep learning methods for fraud detection, valued for their feature extraction and classification capabilities. However, their use is limited compared to more advanced methods such as CNNs, which demonstrate greater accuracy and scalability in handling fraud detection tasks \cite{Alarfaj2022}.

\subsubsection{Machine Learning Models}
The machine learning models can be a baseline or part of a hybrid model: Traditional machine learning algorithms serve as essential benchmarks for financial fraud detection  \cite{Cherkaoui2024}.  Logistic Regression (LR) offers simplicity and interpretability, ideal for binary classification. LinearSVC efficiently handles linear data, while KNN detects anomalies based on proximity, though it lacks scalability for large datasets.

Ensemble methods such as Random Forests and Gradient Boosting (XGBoost, LightGBM) improve accuracy by combining multiple models. Random Forests resist overfitting, while Gradient Boosting handles imbalanced datasets effectively. Adaptive Boosting (AdaBoost) strengthens weak classifiers iteratively for fraud-specific challenges.

Support Vector Machines (SVMs) excel in high-dimensional spaces, identifying outliers effectively. Decision Trees, though prone to overfitting, remain interpretable and foundational for ensemble models.

While less adaptive than deep learning, these algorithms provide valuable benchmarks, offering insights into structured data performance and guiding advancements in fraud detection systems.

\subsubsection{Hybrid Models}
Hybrid models  integrate complementary strengths of different algorithms to address complex tasks like financial fraud detection. 
\textbf{Adaptive Sampling and Aggregation-Based Graph Neural Network (ASA-GNN)} enhances traditional GNN frameworks by integrating adaptive sampling and entropy-based aggregation, which address the major limitations of standard GNNs in handling fraud detection. By focusing on relevant neighbors and combating oversmoothing, ASA-GNN provides a robust and scalable solution for identifying complex fraud patterns in graph-structured data \cite{Tian2024}.

\textbf{Reinforcement Learning with Deep Q-Network (RDQN)} integrates deep learning (DNN) and reinforcement learning(Q learning). By leveraging Rough Set Theory for feature reduction and employing reinforcement learning, the RDQN model achieves faster processing and higher accuracy, making it a scalable and effective solution for real-world fraud detection problems. RDQN outperforms traditional models like SVM, ANN, and DT, as well as hybrid models such as IFDTC4.5, SAE-GAN, and CNN-SVM-KNN \cite{Tekkali2023}. 

\textbf{Transformer-LOF-Random Forest} model uniquely combines the Transformer’s advanced feature extraction, LOF (local outlier factor) ’s local anomaly detection, and Random Forest’s ensemble learning to effectively detect complex and rare fraudulent patterns. It surpasses state-of-the-art models such as XGBoost, LightGBM, and LSTM by addressing data imbalance, reducing false positives and false negatives, and adapting to emerging fraud techniques \cite{Miao2024}.

\textbf{ResNeXt-embedded Gated Recurrent Unit (RXT-J)} integrates the feature extraction capabilities of ResNeXt and the sequential learning strengths of Gated Recurrent Units (GRU). RXT-J model significantly outperforms existing models, including BERT (Transformer), ANN, and logistic regression \cite{Almazroi2023}.

\textbf{CatBoost-Deep Neural Networks combines CatBoost and Deep Neural Networks (DNN)} to leverage their respective strengths. CatBoost excels at handling categorical features, imbalanced datasets, and complex relationships in structured data \cite{Lokanan2024}. while DNN focuses on learning patterns in raw features and adapting to sparse data conditions. The hybrid model significantly outperforms others such as random forests and ensemble methods such as LSTM-based AdaBoost \cite{Nguyen2022}.

\textbf{Autoencoder-LSTM} is a combination of two deep learning models: Autoencoder and LSTM. Autoencoder performs dimensionality reduction while retaining key features and removing noise. LSTM network models temporal dependencies and classifies transactions as fraudulent or legitimate. The combined model demonstrates superior performance over traditional machine learning methods and standalone LSTM models \cite{Zioviris2024}.

\subsubsection{Trends in Model Usage}
\begin{figure}[!h]
\centering
\includegraphics[width=0.85\textwidth]{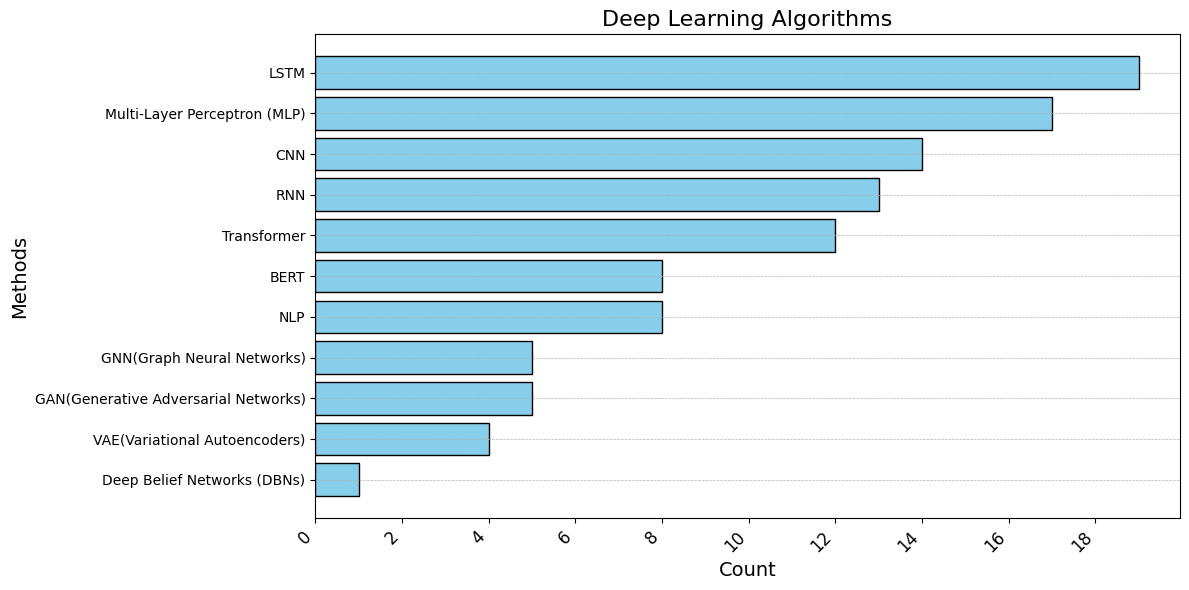}
\caption{Distribution of Deep Learning Techniques Applied to Financial Fraud Detection} \label{dlcount}
\end{figure}

We observe the overall pattern of frequency of application of deep learning models in Figure ~\ref{dlcount}. LSTM, MLP, CNN and RNN are most widely used. NLP methods such as Bidirectional Encoder Representations from Transformer (BERT) model  are moderately applied for textual dataset. Additionally, more specialized methods have been applied to detect fraud, including GNNs, GANs, VAEs. As fraud detection datasets are highly imbalanced, while GANs are moderately applied, their potential in modeling relationships and generating synthetic datasets for fraud detection could drive more research in these areas. 

\begin{figure}[!h]
\centering
\includegraphics[width=0.85\textwidth]{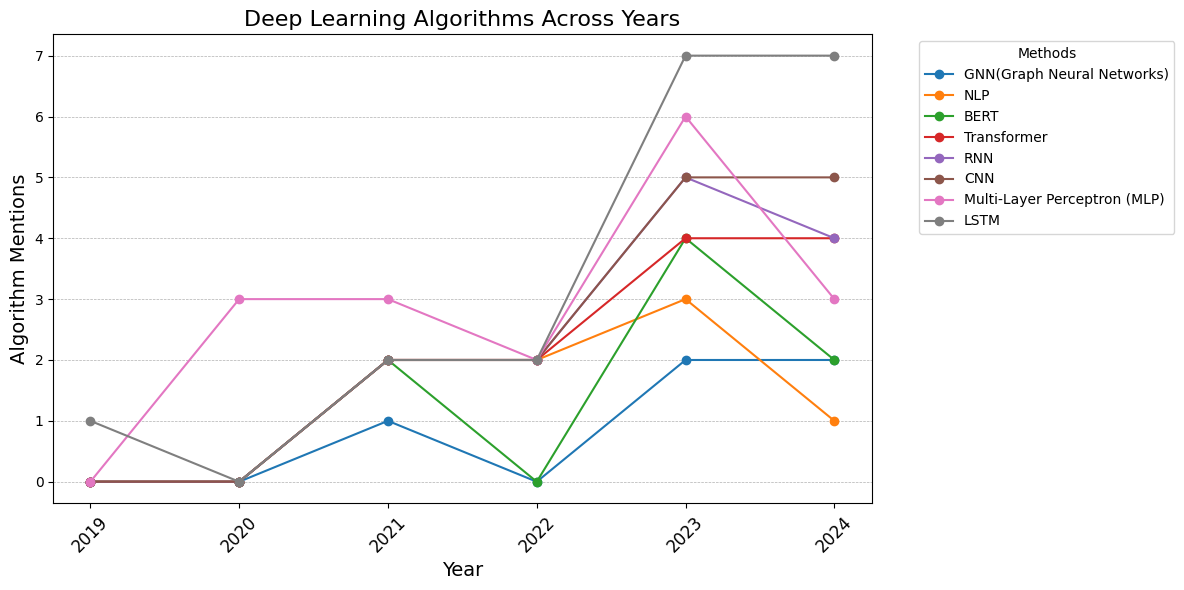}
\caption{Yearly Trends of Deep Learning Algorithm Application in Fraud Detection from 2019 to 2024. } \label{dlcount-year}
\end{figure}

\textbf{Baseline methods.} This study analyzes the yearly trend of deep learning algorithms in fraud detection. As shown in Figure ~\ref{dlcount-year}, overall the variability of deep learning models has increased over years. LSTM has the most significant and sustained growth over the years, culminating in a sharp increase from 2022 to 2024. This trend could be driven by the sequential nature of fraud datasets. MLP and CNNs model maintain a steady trend. These models are versatile and effective in learning complex relationships between features in financial datasets.
\begin{figure}[!h]
\centering
\includegraphics[width=0.85\textwidth]{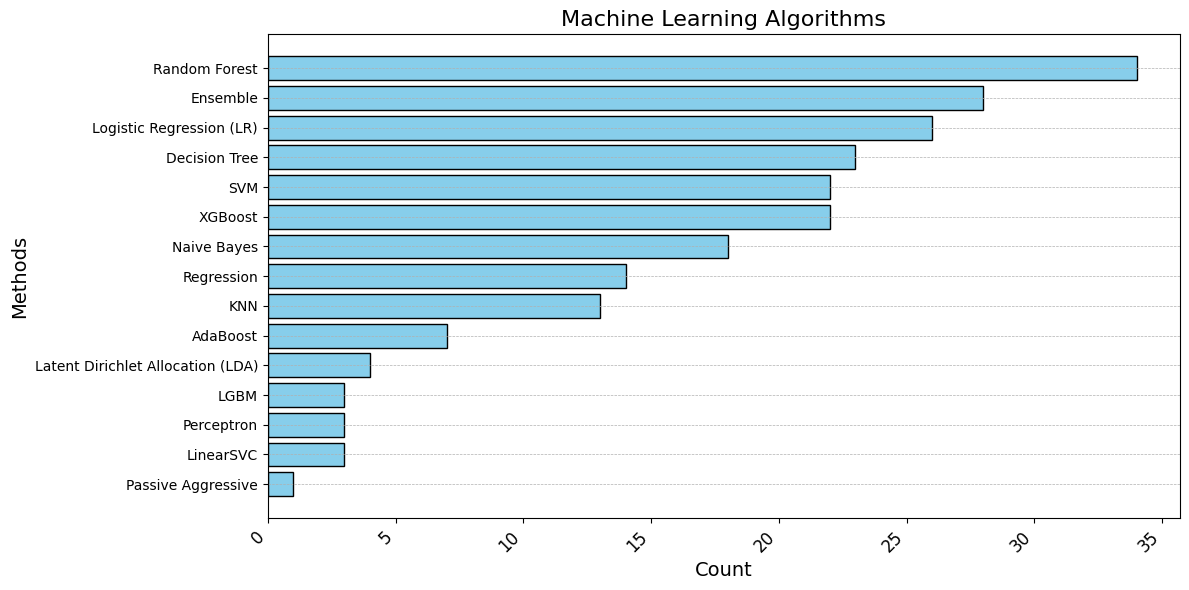}
\caption{Distribution of Machine Learning Techniques Applied to Financial Fraud Detection} \label{mlcount}
\end{figure}

\begin{figure}[!h]
\centering
\includegraphics[width=0.85\textwidth]{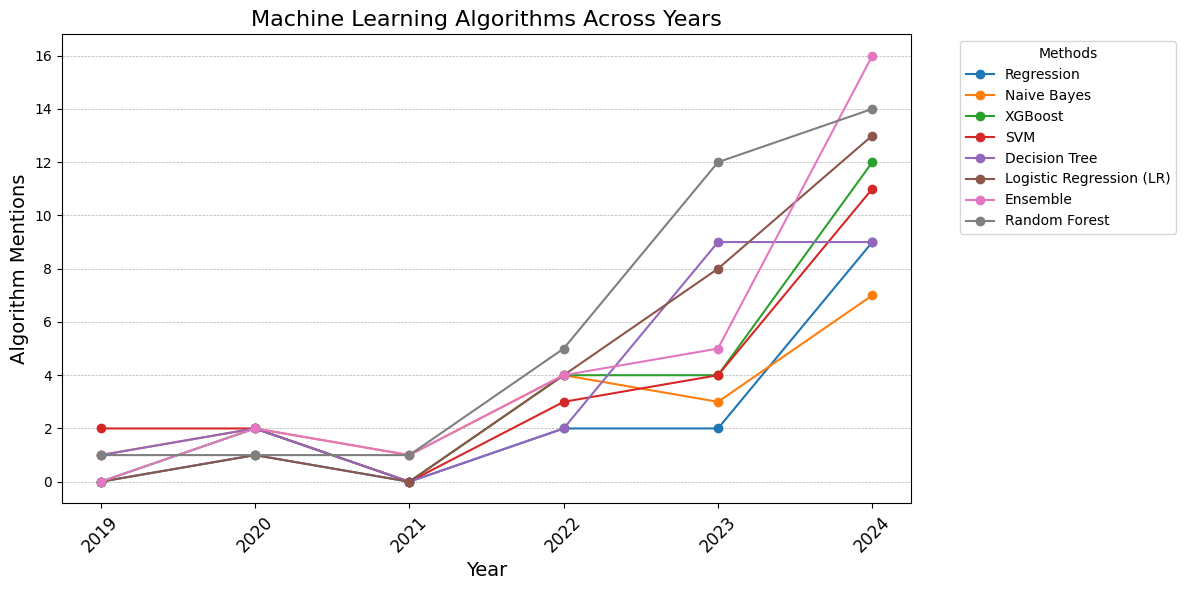}
\caption{Yearly Trends of Machine Learning Algorithm Application in Fraud Detection from 2019 to 2024.} \label{mlcount-year}
\end{figure}

\begin{figure}[!h]
\centering
\includegraphics[width=0.88\textwidth]{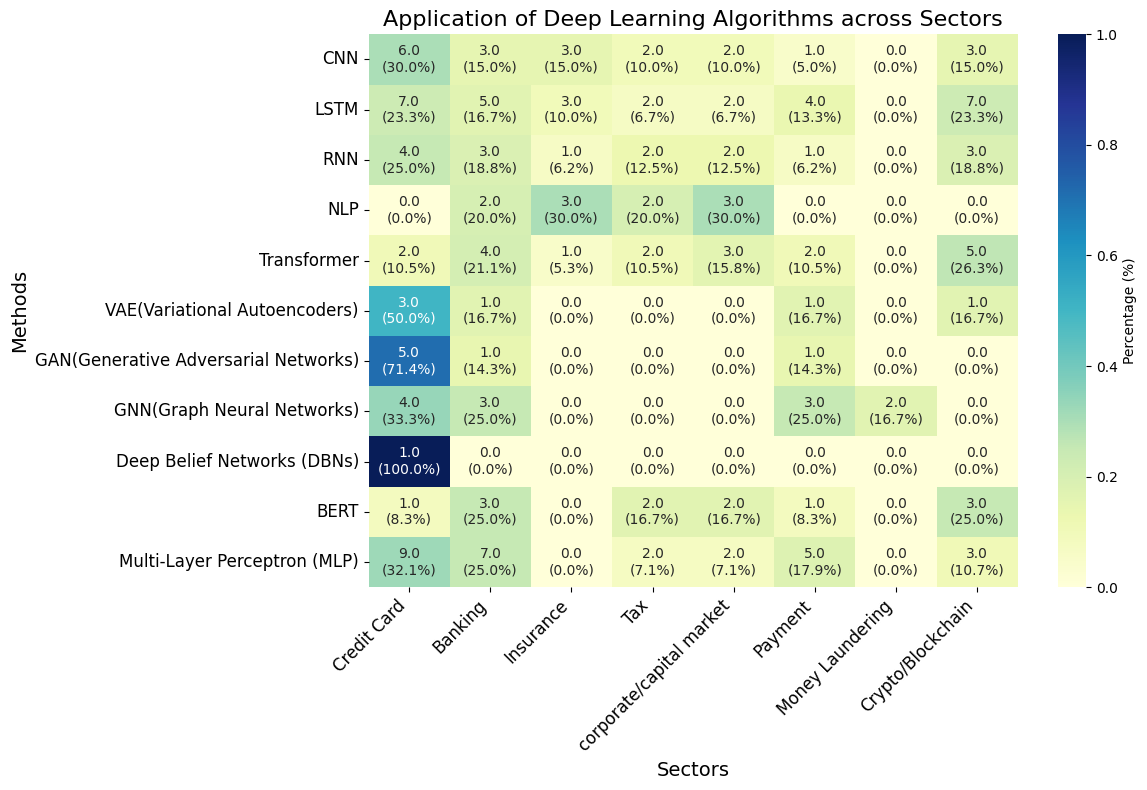}
\caption{Application of Deep Learning Methods in Financial Fraud Detection Across Sectors. } \label{heatmap}
\end{figure}

\textbf{Deep learning methods across sectors.} Figure~\ref{heatmap} demonstrates the distribution and frequency of various deep learning algorithms applied to financial fraud detection across different financial sectors. Both the credit card and banking sectors have significant applications of a wide range of deep learning techniques. Particularly, MLP, LSTM, and CNN have been more commonly used. LSTM and GNN show significant application, likely due to the sequential and graph-structured nature of blockchain data, which requires methods that can model relationships between transactions. Sectors such as Tax and Money Laundering show minimal application of deep learning techniques.

\begin{figure}[!h]
\centering
\includegraphics[width=0.85\textwidth]{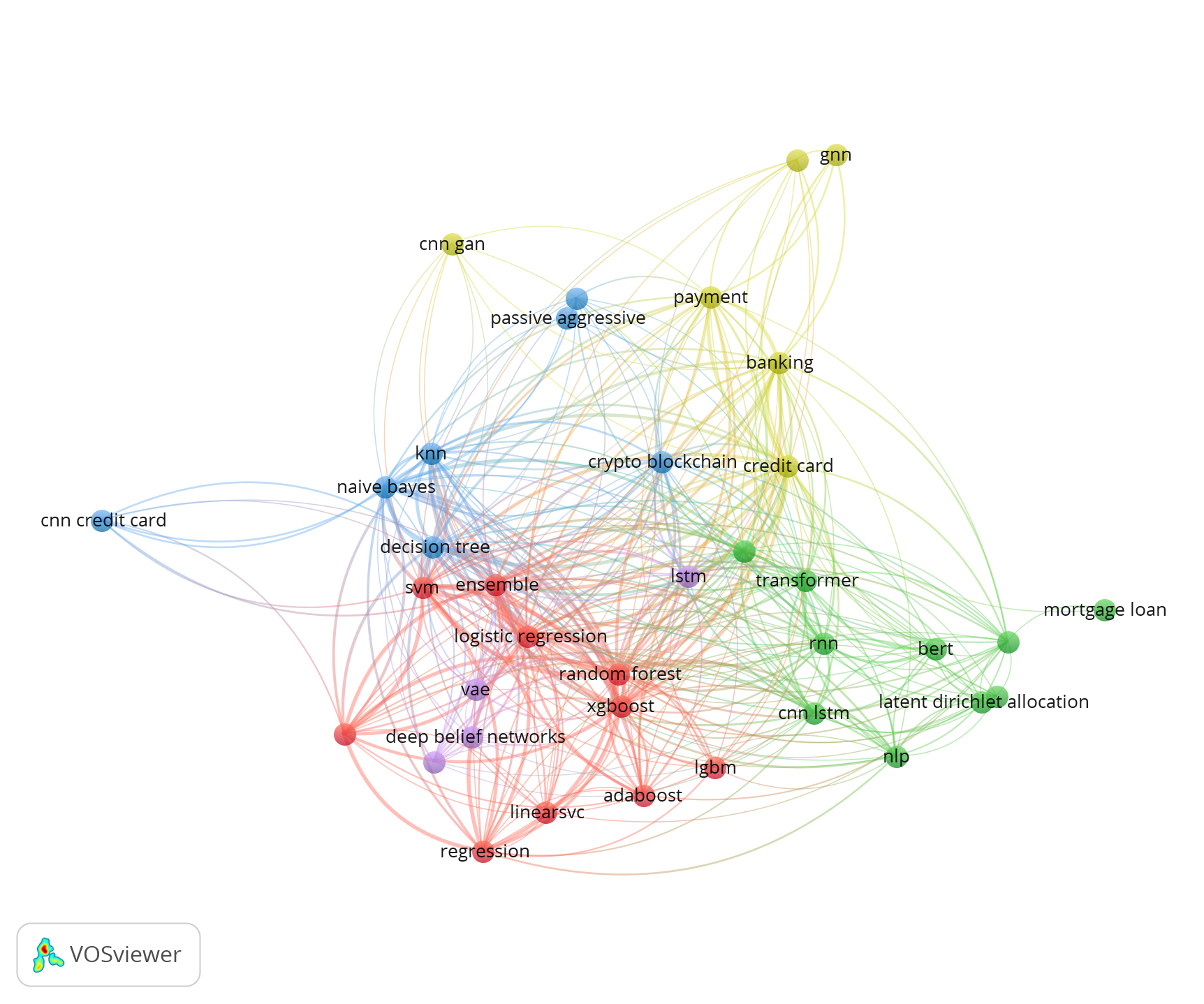}
\caption{Keyword Co-occurrence Network Graph.} \label{network}
\end{figure}

To explore the interconnection between various methods and domains, this study conduct a semantic network analysis in Figure ~\ref{network}: 

Cluster 1 (colored in red) has 10 keywords related to machine learning models, including Random Forest, Logistic Regression, SVM, Ensemble, Decision Tree, etc. Random forest (with 32 links and 234 link strength) and logistic regression (with 31 links and 191 link strength) are the two most relevant keywords. These two traditional machine learning models are most frequently connected with keywords representing deep learning, indicating the models are often used as baselines or hybrid models. 

Cluster 2 (colored in green) has 10 keywords related to deep learning models, including Transformer, BERT, RNN, etc. MLP (with 27 links and 103 link strength) is the most relevant keyword. It is well connected with CNN, LSTM, SVM and  Random Forest. 

Cluster 3 (colored in blue) has 7 keywords related to lightweight models, including Naive Bayes, KNN, decision trees. This smaller cluster represents lightweight models used for simple datasets. 

Cluster 4 (colored in yellow) has 6 keywords related to both financial sectors and deep learning models. Credit card is the most relevant node with 30 links and 160 link strength. It also includes deep learning algorithms such as CNN and GAN, indicating the popular trend of CNN and GNN in fraud detection, especially in credit card and banking.

Cluster 5 (colored in purple) has 4 keywords related to more specialized deep learning models such as GAN, VAE. It shows an emerging trend of applying GAN for generating synthetic data to address imbalanced datasets.

In summary, classical ML methods (e.g., Random Forest, SVM) remain foundational for structured data fraud detection. Techniques like LSTM, Transformers, and BERT show increasing adoption for analyzing sequential and text-based fraud data. GNNs and GANs are gaining traction in specialized fraud domains like blockchain and synthetic data generation. Different algorithms are tailored to the needs of specific financial sectors (e.g., CNN for credit card fraud, GNN for blockchain).

\subsection{Research Question 4}
\textbf{What trends can be observed in the benchmarks and evaluation metrics used to assess the effectiveness of deep learning models across different financial sectors?}\\

Traditional metrics face limitations in handling imbalanced datasets where fraud cases are rare. Metrics like \textbf{Accuracy} often mislead, favoring the majority class and overlooking fraud. \textbf{Precision} minimizes false positives but may reduce \textbf{Recall}, which identifies actual fraud cases. High Recall, while critical, can increase the \textbf{False Positive Rate (FPR)} and operational costs. The \textbf{F1 Score} balances Precision and Recall but does not address economic impacts.

Metrics such as \textbf{False Negative Rate (FNR)} and \textbf{FPR} highlight specific challenges, like losses from undetected fraud or inefficiencies from false alarms. \textbf{AUC-PR} (Precision-Recall Curve) is more relevant than \textbf{AUC-ROC} for imbalanced datasets, focusing on fraud detection effectiveness. The metrics and their formulas are shown in Table \ref{tab:fraud_metrics} below.

Economic metrics, such as the \textbf{Cost of False Positives} and \textbf{Cost of False Negatives}, assess operational and financial impacts, helping organizations optimize detection strategies. For instance, \textbf{Precision} is prioritized in cryptocurrency fraud to minimize compliance costs, while \textbf{Recall} is critical in tax fraud to prevent revenue losses.

In conclusion, effective fraud detection increasingly relies on tailored metrics that address imbalanced datasets, operational costs, and sector-specific needs, complementing foundational metrics like Accuracy and F1 Score with domain-specific benchmarks. 

\begin{table}[h!]
\centering
\resizebox{\textwidth}{!}{ % Resize the table to fit the page width
\begin{tabular}{|>{\centering\arraybackslash}m{3cm}|>{\centering\arraybackslash}m{5cm}|>{\centering\arraybackslash}m{7cm}|}
\hline
\textbf{Metrics Name} & \textbf{Formula} & \textbf{Description} \\
\hline
\textbf{Accuracy} & \(\frac{TP + TN}{TP + TN + FP + FN}\) & Measures the proportion of correctly classified instances (both fraud and non-fraud) out of the total instances. High accuracy is not always reliable in fraud detection due to class imbalance. \\
\hline
\textbf{Precision} & \(\frac{TP}{TP + FP}\) & Measures the proportion of correctly predicted fraud cases out of all predicted fraud cases. High precision indicates fewer false positives (legitimate transactions flagged as fraud). \\
\hline
\textbf{Recall (Sensitivity)} & \(\frac{TP}{TP + FN}\) & Measures the proportion of actual fraud cases correctly identified by the model. High recall indicates fewer false negatives (fraudulent transactions missed by the model). \\
\hline
\textbf{F1 Score} & \(2 \times \frac{Precision \times Recall}{Precision + Recall}\) & The harmonic mean of precision and recall. It balances precision and recall, making it useful when both false positives and false negatives are costly. \\
\hline
\textbf{AUC-ROC} & Area under the ROC curve (TPR vs FPR) & TPR (True Positive Rate) is plotted against FPR (False Positive Rate) at various thresholds. AUC-ROC measures the model's ability to distinguish between fraud and non-fraud cases. Higher AUC-ROC indicates better performance. \\
\hline
\textbf{AUC-PR} & Area under the Precision-Recall curve (Precision vs Recall) & Precision is plotted against Recall at various thresholds. AUC-PR is especially useful for imbalanced datasets (common in fraud detection) as it focuses on the performance of the positive class (fraud). Higher AUC-PR indicates better performance. \\
\hline
\textbf{False Positive Rate (FPR)} & \(\frac{FP}{FP + TN}\) & Measures the proportion of legitimate transactions incorrectly flagged as fraud. Lower FPR is desirable to reduce customer inconvenience. \\
\hline
\textbf{False Negative Rate (FNR)} & \(\frac{FN}{FN + TP}\) & Measures the proportion of fraudulent transactions missed by the model. Lower FNR is critical in fraud detection to minimize financial losses. \\
\hline
\textbf{Cost of False Positives} & \(C_{FP} \times FP\) & Represents the financial or operational cost associated with incorrectly flagging legitimate transactions as fraud (e.g., customer dissatisfaction, manual review costs). \\
\hline
\textbf{Cost of False Negatives} & \(C_{FN} \times FN\) & Represents the financial loss or risk associated with failing to detect fraudulent transactions (e.g., chargebacks, lost revenue). This is typically higher than the cost of false positives in fraud detection. \\
\hline
\end{tabular}
}
\caption{Evaluation Metrics for Fraud Detection}
\label{tab:fraud_metrics}
\end{table}
\subsection{Research Question 5}
\textbf{How have changes in data privacy, anonymization, and regulatory rules influenced the development and application of deep learning models for financial fraud detection?}\\

Of the 57 papers reviewed, 36 of them have publicly available datasets and 24 non-public dataset. 19 of the papers addressed the issue regarding privacy.

\textbf{Principal Component Analysis (PCA)} enhances data privacy in fraud detection by reducing high-dimensional data into lower-dimensional representations, preserving critical information while minimizing information loss \cite{Alarfaj2022}. This transformation anonymizes transactional data, removing identifiable personal information and enabling secure sharing across institutions without exposing sensitive details. Additionally, PCA aids compliance with data protection regulations by mitigating re-identification risks and retaining only essential features needed for fraud detection, reducing unnecessary exposure of sensitive transaction details. However, by altering the original features, PCA can reduce model interpretability, potentially limiting insights into the relationships between variables \cite{Almarshad2023}. Despite this, its ability to balance privacy and performance makes it a valuable tool for secure data processing.

\textbf{Blockchain} enhances data privacy in financial fraud detection through its decentralized and immutable nature, eliminating the need for centralized intermediaries. \cite{Devaguptam2024, Mohammed2024} Sensitive data, such as personal identities and financial transactions, is securely distributed across an encrypted ledger, preventing unauthorized access. Pseudonymity is ensured by cryptographic addresses, further safeguarding personal information \cite{Kapadiya2022}. 

Both the \textbf{General Data Protection Regulation (GDPR)} and the \textbf{California Consumer Privacy Act (CCPA)} significantly impact the application of machine learning in financial fraud detection by enforcing strict data privacy, security, and transparency requirements. GDPR mandates organizations to obtain consent, anonymize sensitive data, and adhere to ethical standards, emphasizing transparency, accountability, and data minimization \cite{Mwangi2024}. Similarly, CCPA grants individuals the right to access, delete, or opt out of data usage, requiring ML models to anonymize or pseudonymize data to protect identities.

A critical GDPR mandate is the "right to explanation," which ensures AI decisions, such as fraud detection outcomes, are interpretable \cite{Kurshan2021}. This presents a challenge for complex models like deep neural networks, prompting the development of explainability techniques to balance compliance and performance. CCPA similarly emphasizes consumer control and data minimization, posing challenges for ML systems reliant on extensive datasets. To comply with both regulations, techniques like data anonymization, federated learning, and secure multiparty computation have become essential. Non-compliance risks penalties under both frameworks, driving organizations to invest in secure, transparent, and privacy-preserving ML systems, which ultimately enhance trust, scalability, and effectiveness in fraud detection.

The \textbf{Cooperative Council for Health Insurance (CCHI)} regulates private health insurance in Saudi Arabia, ensuring fraud prevention, data integrity, and healthcare accessibility. Under Saudi Vision 2030, CCHI mandates detailed records of fraud, fostering transparency and accountability among insurers and providers \cite{Nabrawi2023}.

CCHI’s initiatives, like the National Platform for Health and Insurance Exchange Services (NPHIES), enhance data security and interoperability. Leveraging advanced technologies, including machine learning, the platform improves fraud detection and processing efficiency, aligning with goals to reduce fraud costs and comply with data protection and ethical standards in health insurance.

\section{Discussion}
Fraud detection using advanced techniques is critical in the financial sector due to increasingly sophisticated fraudulent activities. This study examines the use of deep learning across financial domains, highlighting shared techniques and sector-specific challenges. The following sections discuss key applications and advancements in credit card fraud detection, insurance fraud, blockchain integration, and banking systems.

\subsection{Credit Card}
Credit card fraud detection has become a widely researched area with deep learning applications. Due to access to publicly available large-scale datasets, applications of state-of-the-art deep learning and machine learning models have achieved significant success. Techniques such as deep neural networks, and data augmentation could play pivotal roles in overcoming data imbalance and improving model performance. Credit card fraud detection often requires real-time or near-real-time decision-making. Deep learning models can process large volumes of transaction data quickly and accurately, making them ideal for these applications \cite{Paramesha2024, Abbassi2023}.

\subsection{Insurance}
Insurance fraud poses significant challenges due to extensive data exchanges among patients, providers, and insurers, increasing vulnerabilities like data breaches and inaccuracies. Privacy regulations such as HIPAA and GDPR further complicate data sharing by imposing strict safeguards on personal health information \cite{Devaguptam2024}.

Deep learning techniques effectively analyze complex datasets to detect anomalies and predict fraud. Federated learning enables collaborative model training among insurers and providers without exposing sensitive data, ensuring compliance with privacy regulations while enhancing detection efficiency \cite{Kapadiya2022}. Combined with smart automation, these innovations address key challenges in fraud prevention and operational complexity.

\subsection{Blockchain}
Blockchain provides a secure, decentralized ledger that ensures data integrity and privacy, with transparency that enhances fraud detection and auditability, particularly in healthcare insurance  \cite{Kaafarani2024}. Integrating blockchain with machine learning improves fraud detection by leveraging immutable datasets for anomaly detection and prediction \cite{Mavundla2024}.

Permissioned blockchains enable secure data sharing through distributed ledgers, improving accuracy and transparency \cite{Dhieb2020, Aljofey2022}. In the insurance sector, combining blockchain with ML models, such as XGBoost and VFDT, has increased detection accuracy and reduced error rates by 7\% compared to traditional methods \cite{Dhieb2020, Gaikwad2023}. Additionally, smart contracts streamline claims processing, reducing errors and processing time \cite{Aljofey2022, Krishnan2023}. Despite challenges like scalability and compliance, these technologies are highly effective in insurance fraud detection and lay the foundation for secure, efficient systems.

\subsection{Banking \& Payment }
The banking sector faces increasingly sophisticated fraud due to the growth of online banking and diverse payment channels, especially in money laundering, which impacts economies at multiple levels. Traditional rule-based systems often fail to adapt to evolving fraud patterns, leading to high false positive rates. Deep learning models, such as LLMs, analyze complex data to identify anomalies in transactions and spending behaviors, flagging potential fraud effectively \cite{Xu2024, Paramesha2024}. Key challenges include real-time detection, managing large-scale transnational networks, and integrating diverse data sources like Know Your Customer (KYC) profiles. Addressing these issues requires innovative, scalable solutions to combat financial fraud efficiently \cite{Usman2023}.

\section{Limitation and Future Direction}
This study systematically reviews advancements in deep learning for financial fraud detection but has limitations in its approach. First, the selection of studies between 2019 and 2024 excludes earlier foundational work that may provide additional context. Second, while focusing on publicly available literature ensures transparency, it leaves out proprietary models and industry-specific practices that might offer innovative insights. Third, inconsistencies in data processing and evaluation frameworks across the reviewed studies make cross-comparison of findings challenging. Additionally, the reliance on English-language publications restricts insights from regions where other languages predominate, potentially missing localized advancements or applications.

Future research should prioritize standardization across the Data Science lifecycle to enhance reproducibility and generalizability in financial fraud detection. Key areas of focus include:
\begin{itemize}
    \item \textbf{Data Preparation}: Develop unified preprocessing pipelines that address imbalanced datasets through advanced techniques like GANs, SMOTE, and feature scaling to improve model reliability across studies.
    \item \textbf{Model Development}: Encourage collaboration between academia and industry to bridge research and real-world implementation, incorporating state-of-the-art techniques like explainable AI and automated parameter tuning for scalable deployment.
    \item \textbf{Model Evaluation}: Establish comprehensive benchmarks tailored to specific financial domains, integrating economic and operational metrics to better reflect real-world costs and impacts.
    \item \textbf{Operational Deployment}: Focus on integrating deep learning models with existing systems, emphasizing robust APIs, real-time fraud detection, and compliance with data privacy regulations.
    \item \textbf{Monitoring and Maintenance}: Promote adaptive frameworks that allow ongoing model tuning and retraining, ensuring systems evolve alongside emerging fraud techniques and regulatory changes.
\end{itemize}
By addressing these directions, future studies can provide more actionable, scalable, and compliant solutions, contributing to the ongoing advancement of deep learning applications in fraud detection.

\section{Conclusion}
The findings from this systematic literature review underscore the transformative role of deep learning in financial fraud detection. By analyzing recent advancements, it becomes clear that deep learning models, including CNNs, LSTMs, transformers, and ensemble techniques, have significantly enhanced the ability to detect complex fraud patterns across diverse financial sectors. These models, combined with robust preprocessing and feature engineering techniques, address key challenges such as imbalanced datasets and operational scalability, leading to more accurate and efficient fraud detection systems.

Moreover, the integration of privacy-preserving methods, such as blockchain, PCA, and compliance with global regulations like GDPR and CCPA, ensures that these advancements are aligned with ethical and legal standards. These measures not only protect sensitive financial and personal data but also build trust among stakeholders and foster broader adoption of advanced fraud detection solutions. The exploration of automation techniques, including parameter tuning and subsampling, has further accelerated time-to-detection, making real-time fraud mitigation feasible in high-stakes environments.

As fraud detection systems evolve, future research should focus on enhancing model interpretability, addressing emerging fraud schemes, and improving cross-industry collaboration through federated learning and secure data sharing frameworks. This will ensure that deep learning systems remain resilient and adaptable in an ever-changing financial landscape. The insights from this study serve as a foundation for advancing both the technical and operational aspects of fraud detection, fostering a more secure and trustworthy financial ecosystem.

% \printbibliography  % Use this instead of \bibliography{}

% \nocite{*} % Include all references from the .bib file
% % \begin{thebibliography}{99}
% \bibliographystyle{apalike}  % You can use other styles like ieee, apa, etc.
% \bibliography{references}  % This refers to your .bib file (without the .bib extension)

% \section{Disclosure Statement}
% No potential conflict of interest was reported by the author(s).

% \section{Author Details}
% \input{author_details}

% \section{Author Contributions}
% \input{author_contributions}

% \section{Funding}
% N/A

% \end{thebibliography}

\end{document}